\documentclass[format=sigconf]{acmart}
\usepackage{graphicx} 
\usepackage{hyperref}
\usepackage{acronym}
\usepackage[frozencache,cachedir=.]{minted}

\setminted[yaml]{breaklines}

\usepackage{framed}

\AtBeginDocument{%
  }

\setcopyright{acmlicensed}
\copyrightyear{2025}
\acmYear{2025}
\acmDOI{XXXXXXX.XXXXXXX}
\acmConference[GECCO '25]{The Genetic and Evolutionary Computation Conference 2025}{July 14--18, 2025}{Málaga, Spain}

\acrodef{llm}[LLM]{Large Language Model}
\acrodef{ge}[GE]{Guided Evolution}
\acrodef{crp}[CRP]{Character Role Play}
\acrodef{eot}[EoT]{Evolution of Thought}
\acrodef{ea}[EA]{Evolutionary Algorithm}
\acrodef{yolo}[YOLO]{You Only Look Once}
\acrodef{nas}[NAS]{Neural Architecture Search}
\acrodef{rag}[RAG]{Retrieval Augmented Generation}
\begin{document}
\title{LLM-Guided Evolution: An Autonomous Model Optimization for Object Detection}

\author{YiMing Yu}
\email{YiMing.Yu@gtri.gatech.edu}
\orcid{0000-0002-3467-122X}
\affiliation{%
  \institution{ Georgia Tech Research Institute}
  \city{Atlanta}
  \state{Georgia}
  \country{USA}
}

\author{Jason Zutty}
\email{Jason.Zutty@gtri.gatech.edu}
\orcid{0000-0001-7977-1454}
\affiliation{%
  \institution{ Georgia Tech Research Institute}
  \city{Atlanta}
  \state{Georgia}
  \country{USA}
}


\begin{abstract}
In machine learning, Neural Architecture Search (NAS) requires domain knowledge of model design and a large amount of trial-and-error to achieve promising performance. Meanwhile, evolutionary algorithms have traditionally relied on fixed rules and pre-defined building blocks. The Large Language Model (LLM)-Guided Evolution (GE) framework transformed this approach by incorporating LLMs to directly modify model source code for image classification algorithms on CIFAR data and intelligently guide mutations and crossovers. A key element of LLM-GE is the "Evolution of Thought" (EoT) technique, which establishes feedback loops, allowing LLMs to refine their decisions iteratively based on how previous operations performed. In this study, we perform NAS for object detection by improving LLM-GE to modify the architecture of You Only Look Once (YOLO) models to enhance performance on the KITTI dataset. Our approach intelligently adjusts the design and settings of YOLO to find the optimal algorithms against objective such as detection accuracy and speed. We show that LLM-GE produced variants with significant performance improvements, such as an increase in Mean Average Precision from 92.5\% to 94.5\%. This result highlights the flexibility and effectiveness of LLM-GE on real-world challenges, offering a novel paradigm for automated machine learning that combines LLM-driven reasoning with evolutionary strategies.
\end{abstract}



\maketitle

\section{Introduction}

\ac{nas} and Object Detection are two interrelated challenges in machine learning, each with significant hurdles. Object detection requires both precise localization and accurate classification, making model design inherently complex. Traditional handcrafted architectures often struggle with generalization across datasets, hardware constraints, and real-time processing needs. \ac{nas} addresses this by automating the search for optimal architectures, efficiently exploring vast design spaces to tailor models for specific detection tasks. However, \ac{nas} itself is computationally expensive, requiring advanced search strategies like reinforcement learning, evolutionary algorithms, or gradient-based methods to balance accuracy and efficiency. The integration of LLMs into \ac{nas} for object detection is crucial for developing scalable, high-performance models capable of handling challenges such as scale variation, occlusion, and deployment on resource-constrained devices. Solving these challenges will enable more robust, efficient, and widely applicable object detection systems, from autonomous vehicles to edge computing.

In 2024 Morris et al. \cite{morris2024llm} introduced a novel technique they coined \ac{llm}-\ac{ge}. They demonstrated its utility by pairing the \ac{llm} Mixtral \cite{jiang2024mixtral} with the ExquisiteNetV2 \cite{zhou2021novel} classifier for the CIFAR-10 dataset and evolved image classifiers with higher accuracy and fewer parameters.

This paper builds upon the research thus far utilizing \ac{llm}-\ac{ge} on the problem of object detection for the KITTI dataset \cite{doi:10.1177/0278364913491297}. This is the first known application of the integration of an \ac{llm} with an evolutionary algorithm for the problem of object detection. This paper introduces a novel approach of using an \ac{llm} to intelligently modify neural network architectures through their module, layer, and hyperparameter descriptions in YAML configuration files, which is a major step forward in automated machine learning and \ac{nas}. This paper demonstrates the powerful capabilities of \ac{llm}-\ac{ge} with its \ac{crp} and \ac{eot} to outperform human state-of-the-art algorithms in the \ac{yolo}  \cite{redmon2016you} family on key objectives of mAP@50, mAP@50-95, precision, and recall. 

\section{Background}
\subsection{Automated Machine Learning}
The development of machine learning traditionally requires expertise in multiple areas, from data preprocessing to feature engineering, model selection, hyperparameter tuning, and deployment. As the demand for AI-driven solutions grows across industries and the complexity of machine learning tasks increases, Automated Machine Learning (AutoML) has emerged as a solution to address these challenges. AutoML automates the end-to-end process of developing machine learning models, enabling non-experts to build high-performance models efficiently while reducing the need for manual intervention. 

AutoML consists of several key components that automate different stages of the machine learning pipeline. Data preprocessing and feature engineering involve data cleaning, transformation, and feature extraction, ensuring that datasets are prepared efficiently without manual intervention \cite{hutter2019automl}. Model selection enables AutoML to explore various algorithms to identify the model with the best performance. Hyperparameter tuning alters model parameters and training configurations using techniques such as grid search and genetic algorithms to enhance accuracy and efficiency \cite{bergstra2011hyperparameter}. For deep learning architecture design, \ac{nas} automates the discovery of optimal network architectures. Finally, model evaluation and deployment ensure that selected models are validated using cross-validation, allowing seamless integration into real-world applications \cite{olson2016tpot}. These components work together to streamline machine learning development, making AI more accessible and scalable across various domains \cite{ledell2020h2o}.

Frequent in the development of AutoML solutions are \ac{ea}-based approaches \cite{olson2016tpot, zutty2015multiple}, which utilize a genetic encoding of machine learning pipeline choices and their hyperparameters.

AutoML has demonstrated its ability to surpass human-designed architectures in diverse domains, from image recognition (NASNet \cite{zoph2017nas}, EfficientNet \cite{tan2019efficientnet}) to structured data analysis (AlphaD3M \cite{drori2019alphad3m}).

The overall goal of research in autoML is to discover higher-performing algorithms with fewer evaluations of candidate solutions or less total computational time. This paper considers an emerging new subfield of autoML which involves the usage of \acp{llm}, described in Section \ref{sec:EC+LLM}.

\subsection{Evolutionary Computation integrated with \acp{llm}}
\label{sec:EC+LLM}

The recent advancements in \ac{llm} capabilities have jump-started a new intersection between evolutionary computation and the employment of \acp{llm} for automated machine learning \cite{wu2024evolutionary}. \acp{llm} excel in code production due to the structure and rules followed in each language. 

While traditional AutoML and evolutionary approaches rely on a representation or encoding of a genome for each algorithm, an evolutionary algorithm paired with an \ac{llm} is able to operate directly on code \cite{hemberg2024evolving}. Such an \ac{ea} still retains its bio-inspired metaphor, where evaluation, selection, crossover, and mutation are in play. In this new form, evaluation and selection are handled using standard \ac{ea} approaches, but the operators of crossover and mutation utilize \acp{llm}, where each operation involves prompting an \ac{llm} with the code that is the individual. 

Recently, this type of approach has been demonstrated several times on common benchmark problems such as image classification with the CIFAR-10 dataset \cite{morris2024llm, 10.1145/3638529.3654017} or MNIST \cite{chen2302evoprompting} and the traveling salesman problem \cite{liu2023algorithm}. An open question this paper seeks to answer is does this approach work well with less frequently published datasets and problems, where the \ac{llm} may not have been trained with as much information?

Across open literature, a variety of \acp{llm} have been used for integration with \acp{ea} including: CodeGen-6B, Mixtral 8x7B, GPT-3.5 Turbo, Llama-2-70B-Instruct, and PaLM-62B. However, few papers compare the performance of \acp{llm} within optimizations or blend their results.


\subsection{Neural Architecture Search for Object Detection}

Handcrafted neutral architecture requires domain knowledge of model design and large amounts of trail-and-error experiments to achieve promising performance for classification, detection, and segmentation tasking. As \ac{nas} demonstrated its efficacy in designing SoTA image classification models  \cite{zoph2017nas, tan2019efficientnet}, the \ac{nas} emerged to design detection backbones for object detection initially \cite{chen2019detnas}.

A state-of-the-art object detection system typically consists of a backbone, a feature fusion neck, a region proposal network (RPN), and an RCNN head. In one-stage detection models, such as \ac{yolo}, the detection pipeline consists of only a backbone and a head. \ac{nas} has emerged as a powerful tool to automate the design of detection architectures under given constraints, often guided by principles such as maximum entropy. The MAE-DET model was proposed to automatically design detection backbones using the Maximum Entropy Principle, eliminating the need for network training while still achieving state-of-the-art performance \cite{sun2022maedetrevisitingmaximumentropy}. Similarly, DAMO-\ac{yolo} was developed to optimize the backbone, neck, and head, resulting in a new model that surpasses previous \ac{yolo} series models on the COCO dataset \cite{xu2022damo}.

In addition, EAutoDet was introduced to design both the backbone and Feature Pyramid Network (FPN) by constructing a supernet that jointly optimizes these modules using a differentiable \ac{nas} method \cite{wang2022eautodet}. Another approach, Structural-to-Modular \ac{nas} \cite{yao2020sm}, applies a multi-objective search algorithm to explore module combinations for backbone optimization, aiming to find efficient and effective architectures for object detection \cite{yao2020sm}.

While none of the existing \ac{nas} methods for object detection have utilized Large Language Models, in this paper, we introduce a novel LLM-driven \ac{nas} approach. Specifically, we use a LLM-GE to intelligently modify neural network architectures through their module, layer, and hyperparameter descriptions in YAML configuration files, effectively enhancing \ac{yolo}’s performance on the KITTI dataset.

\section{LLM-Guided Evolution Framework}
\label{sec:llm-ge}

The LLM-Guided Evolution framework \cite{morris2024llm} introduces Guided Evolution, which combines LLMs with evolutionary algorithms to automate and enhance \ac{nas}. GE leverages the Evolution of Thought (EoT) feedback loop and Character Role Play (CRP) to intelligently guide both mutations and crossovers in code, making \ac{nas} more efficient and creative. In their study, the authors utilized Mixtral AI's 8x7B Mixture of Experts Model \cite{jiang2024mixtral}, known as Mixtral. The mixtral employs a Mixture of Experts architecture with eight experts, each containing 7 billion parameters. During inference, only a subset of experts is active, which leads to reduced computational overhead and results in efficient processing for tasks requiring rapid inference.



Instead of modifying Python code, our study uses a YAML configuration file for \ac{yolo} as the seed model. To effectively modify a YAML file for \ac{yolo}, an \ac{llm} must possess the following core abilities:

\begin{enumerate}
    \item \textbf{Code Understanding:} Familiarity with YAML syntax and \ac{yolo}-specific configurations.
    \item \textbf{Domain Knowledge:} Understanding of \ac{yolo} architecture and relevant machine learning concepts.
    \item \textbf{Contextual Adaptability:} The ability to adapt configurations to specific use cases.
\end{enumerate}

Additionally, the \acp{llm} must have the following reasoning capabilities:

\begin{enumerate}
    \item \textbf{Logical Reasoning:} To create a valid YAML file that adheres to \ac{yolo}'s structure and to understand how changes affect the model's performance.
    \item \textbf{Problem-Solving:} To suggest changes that optimize performance. Identifying and resolving invalid configurations
    \item \textbf{Creativity:} To propose innovative modifications, such as adding or removing layers, adjusting scaling factors, or implementing new layers to balance performance and efficiency.
    \item \textbf{Generalization:} To apply changes based on abstract user requirements.
\end{enumerate}

This combination of core abilities and reasoning capabilities make it a challenge for the LLMs to effectively modify \ac{yolo} YAML configurations and achieving optimal performance.

\section{Application to YOLO Optimization on KITTI data}
The KITTI benchmark suit is a widely used real-world computer vision benchmark, providing data for stereo, optical flow, visual odometry, 2D / 3D object detection, and 2D / 3D tracking. For autonomous driving tasks, it provides a comprehensive set of images and corresponding annotations for object detection, scene understanding, and visual odometry. The dataset includes images captured in real-world urban, rural, and highway environments, making it a challenging and representative benchmark for evaluating object detection models. The KIITI object detection dataset contains 7,481 training images and 7,518 test images, with a total of 51,865 annotations for 9 object classes, which are: Car, Pedestrian, Van, Cyclist, Truck, Misc, Tram, Person Sitting, and "Don't care". These objects are labeled with bounding boxes in various sizes. \cite{doi:10.1177/0278364913491297}


In this work, we focus on optimizing the \ac{yolo} object detection model to improve its performance on the KITTI object detection dataset. The \ac{yolo} is a popular real-time detection algorithm due to its single-stage approach and its iterative progression. Since \ac{yolo}v1's inception in 2016 \cite{redmon2016you} it has gone through 11 subsequent iterations to reach \ac{yolo}v12 \cite{tian2025yolov12attentioncentricrealtimeobject}. It is well known for its real-time inference capabilities and the balance between its performance and speed. The model predicts bounding boxes and class probabilities simultaneously with low latency, making it well-suited for tasks in autonomous driving. 

The performance of a \ac{yolo} model can be affected by several factors. In this paper, we will consider the following factors:

\paragraph{Network architecture:}
The network architecture has a significant impact to performance. Various versions of \ac{yolo} (e.g., \ac{yolo}v3, \ac{yolo}v4, \ac{yolo}v5 ...) impact performance due to significant architecture innovations in their backbone networks and head networks. These innovations include the introduction of a Feature Pyramid Networks (FPN) in \ac{yolo}v3, the introduction of a CSPDarknet backbone in \ac{yolo}v5, and the introduction of spatial channel decoupled downsampling and large kernel convolutions in \ac{yolo}v10.\cite{hussain2024yolov5, DBLP:journals/corr/abs-1804-02767, 10473783}

\paragraph{Training Strategies}
In \ac{yolo}'s training framework, data augmentation stands out as a dynamic and practical mechanism. The data augmentation technique includes scale, translation, rotation, shear, random scaling, random erasing, random cropping, and Mosaic transformations. The Mosaic was introduced in \ac{yolo}v4 training to enhance robustness. In addition, hyperparameter tuning for learning rate, weight decay, momentum, and optimizer selection can significantly impact training and generalization.

Designing a network architecture and optimizing its parameters for training manually is a time-intensive process, requiring both domain knowledge and trial-and-error experimentation. By leveraging the LLM-Guided Evolution, we aim to automate this optimization process. During the evolution, the LLM intelligently refines \ac{yolo}'s configuration and hyperparameters through guided mutations and crossovers. This process enables the model to adapt to the specific characteristics of the KITTI dataset for the desired objectives.

This study integrates The LLM-Guided Evolution was with the Ultralytics \ac{yolo} \cite{ultralytics_yolo} repository. Ultralytics \ac{yolo} is a state-of-the-art open-source framework, representing the latest advancement in the renowned \ac{yolo} series for real-time object detection and image segmentation. It includes all previous versions of \ac{yolo} and enhances performance, flexibility, and efficiency by integrating advanced features and continuous improvements from the latest \ac{yolo} architecture developments and publications. In the Ultralytics \ac{yolo} repository, the models are defined using model configuration files represented with YAML, where each YAML file defines a unique model architecture. In general, the YAML file contains essential details to construct a model, such as the number of layers, types of layers, the upper connection of the layers, activation function, input argument of the given layers, and other settings. It includes the following components:

\begin{itemize}
    \item \textbf{Backbone}: Defines the feature extraction network, such as Darknet53 for \ac{yolo}v3 or other backbone architectures.
    \item \textbf{Head}: Defines additional layers for handling specific task, such as objective detection
    \item \textbf{Other parameters}: Defines model parameters, such as input size or multiples for depth and width. It can also contain hyperparameters for training such as learning rate and weight decay. 
\end{itemize}

\begin{listing}[ht]
\begin{minted}[fontsize=\scriptsize]{yaml}
# --Block--
# Parameters
nc: 80 # number of classes
depth_multiple: 1.0 # model depth multiple
width_multiple: 1.0 # layer channel multiple

# --Block--
# backbone
backbone:
  # [from, number, module, args]
  - [-1, 1, Conv, [32, 3, 1]] # 0
  - [-1, 1, Conv, [64, 3, 2]] # 1-P1/2
  - [-1, 1, Bottleneck, [64]]
  - [-1, 1, Conv, [128, 3, 2]] # 3-P2/4
  - [-1, 2, Bottleneck, [128]]
  - [-1, 1, Conv, [256, 3, 2]] # 5-P3/8
  - [-1, 8, Bottleneck, [256]]
  - [-1, 1, Conv, [512, 3, 2]] # 7-P4/16
  - [-1, 8, Bottleneck, [512]]
  - [-1, 1, Conv, [1024, 3, 2]] # 9-P5/32
  - [-1, 4, Bottleneck, [1024]] # 10
  
# --Block--
# head
head:
  - [-1, 1, Bottleneck, [1024, False]]
  - [-1, 1, Conv, [512, 1, 1]]
  - [-1, 1, Conv, [1024, 3, 1]]
  - [-1, 1, Conv, [512, 1, 1]]
  - [-1, 1, Conv, [1024, 3, 1]] # 15 (P5/32-large)

  - [-2, 1, Conv, [256, 1, 1]]
  - [-1, 1, nn.Upsample, [None, 2, "nearest"]]
  - [[-1, 8], 1, Concat, [1]] # cat backbone P4
  - [-1, 1, Bottleneck, [512, False]]
  - [-1, 1, Bottleneck, [512, False]]
  - [-1, 1, Conv, [256, 1, 1]]
  - [-1, 1, Conv, [512, 3, 1]] # 22 (P4/16-medium)

  - [-2, 1, Conv, [128, 1, 1]]
  - [-1, 1, nn.Upsample, [None, 2, "nearest"]]
  - [[-1, 6], 1, Concat, [1]] # cat backbone P3
  - [-1, 1, Bottleneck, [256, False]]
  - [-1, 2, Bottleneck, [256, False]] # 27 (P3/8-small)
  - [[27, 22, 15], 1, Detect, [nc]] # Detect(P3, P4, P5)
\end{minted}
    \caption{YAML configuration of seed model consists of parameters block, backbone block, and head block.}
    \label{lst:seed_model}
\end{listing}
In our study, the YAML file is divided into blocks according to the backbone, head, and parameters. These are three natural segmentation points for the configuration of a \ac{yolo} architecture. Listing~\ref{lst:seed_model} shows the YAML file for \ac{yolo}v3, which comprises a parameter block, a backbone block, and a head block. These blocks function analogously to genetic segments in a genome, which can be modified by the LLM-Guided Evolution via mutations and crossovers.

Each layer in Listing~\ref{lst:seed_model} follows the format \textbf{[from, number, module, args]}, where:  
\begin{itemize}
    \item \textbf{from} indicates the index of the previous layer used as input,
    \item \textbf{number} specifies the number of times the module is repeated,
    \item \textbf{module} defines the type of layer (e.g., \textbf{Conv}),
    \item \textbf{args} contains parameters specific to that module, such as kernel size, stride, and number of channels.
\end{itemize}

An ordered set of these individual layers forms either a \textbf{Head} or a \textbf{Backbone}. All of these elements form a search space that the LLM-GE explores to construct new model architectures.

We selected \ac{yolo}v3 as our seed model for Llama-3.3-GE1 due to its high efficiency in real-time applications, making it an ideal candidate for evaluating whether LLMs possess knowledge of advanced layers introduced in later \ac{yolo} versions.

In our study, we use mAP@50, mAP@50-95, precision, recall, total number of parameters, and inference speed as our objectives for GE. The Mean Average Precision (mAP), is a primary metric for object detection, and is computed as the mean of average precision (AP) across all classes. For mAP@50, the AP are computed at an Intersection over Union (IoU) threshold of 0.5. For mAP@50-90, the mean AP averaged over multiple IoU thresholds ranging from 0.5 to 0.95 with increments of 0.5. Precision measures how many of the detected objects are actually correct, while recall measures how many of the actual objects are correctly detected. The total number of parameters determines the storage and memory requirements of the model, impacting its efficiency and deployment feasibility. Inference speed is measured as the average time to process a single image, which is crucial for real-time applications.

Our study demonstrates LLM-Guided Evolution can explore the search space of configurations of \ac{yolo}, generating \ac{yolo} variants with significant improvements in detection performance. 

\section{Results}
In this investigation, our goal is to use LLM-Guided Evolution to generate \ac{yolo} variants that optimize performance for object detection on KITTI data. We focus on objectives relating to algorithm task performance, as well as computational efficiency of the model. 

\subsection{Evaluation of LLM-GEs}
To enable meaningful evolution, \acp{llm} must possess the core abilities and reasoning capabilities outlined in Section~\ref{sec:llm-ge} to propose valid YAML configurations that can be successfully evaluated within the training framework, ultimately leading to performance improvements. For this study, we ran the evolution in two manners of operation. In the first, which we call Llama-3.3-GE1, we seeded our \ac{llm}-\ac{ge} with the YAML configuration file split into the three blocks shown in Listing~\ref{lst:seed_model}. In this manner of operation we seeded the evolution with only the configuration for \ac{yolo}v3. In the second manner of operation, which we call Llama-3.3-GE2, we made the following changes: 
\begin{enumerate}
    \item We modified the representation of the configuration file to be a single block comprising the head, backbone, and hyperparameters.
    \item We modified the prompts to the \ac{llm} to instruct it to modify a specific part of the YAML file, such as the the head, backbone, or parameters.
    \item We included additional seeds into the starting population of the \ac{llm}-\ac{ge} including \ac{yolo}v3-tiny,  \ac{yolo}v9s, \ac{yolo}v10-tiny, and \ac{yolo}v11, in addition to the \ac{yolo}v3 seed used in the first manner of operation
\end{enumerate}
Table~\ref{tab:llm_ge_numbers} summarizes the framework’s performance over 50 generations for Llama-3.3-GE1 and 59 generations for Llama-3.3-GE2. Note that GE1 produced a 40.7\% failure rate, where the YAML files during the evaluation process, while GE2 produced a 51.9\% failure rate. A possible contribution to the increase could be the added variety of seeds, or the more open prompting style to the \ac{llm}.

\begin{figure}[h]
  \centering
  \includegraphics[width=\linewidth]{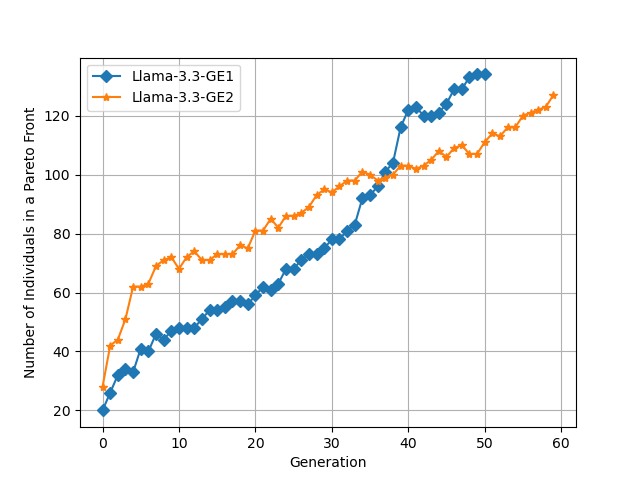}
  \caption{Number of Individuals in the Pareto frontier per generation.}
  \Description{}
  \label{fig:Pareto_frontier_counts}
\end{figure}

Figure~\ref{fig:Pareto_frontier_counts} illustrates the number of individuals on the Pareto frontier over time for each GE. The figure shows that the number of individuals on the Pareto frontiers for both of the GEs grows steadily throughout the generations. Note that the two experiments ran for different numbers of generations: LLM-GE1 stops after 50 generations, while LLM-GE2 stops after 59 generations. The GEs exhibit upward trends, suggesting continued search of the tradeoff space between objectives. Both trends can be characterized by a linear behavior and the trend of Llama-3.3-GE1 has a steeper slope than that of the Llama-3.3-GE2, indicating a faster improvement for the number of individuals in the Pareto Fronts.

\begin{table}
    \centering
 \caption{Performance of LLM-GEs.}
    \begin{tabular}{lccc}
         \toprule
         & Llama-3.3-GE1 & Llama-3.3-GE2 \\
         \toprule
        Total runtime & 38 days & 35 days \\
        Total generation &    50 &    59 \\
        No. of variants &  1891 &  1930   \\
        No. of invalid variants &   769 &   1001   \\
        No. of Pareto frontier  &  135 &  128   \\
         \toprule   
    \end{tabular}
    \label{tab:llm_ge_numbers}
\end{table}

In addition, to compare the overall performance of LLM-Guided Evolution with difference in seedings and content included in prompts, we evaluate the quality of generated solutions using hypervolume under the Pareto frontiers (for minimization of objectives). A smaller hypervolume indicates broader exploration of the Pareto-optimal space. The optimization objectives include the number of parameters, inference speed, precision, and recall,  where \ac{ge} minimizes the number of parameters and inference speed while maximizing precision and recall.

\begin{figure}[h]
  \centering
  \includegraphics[width=\linewidth]{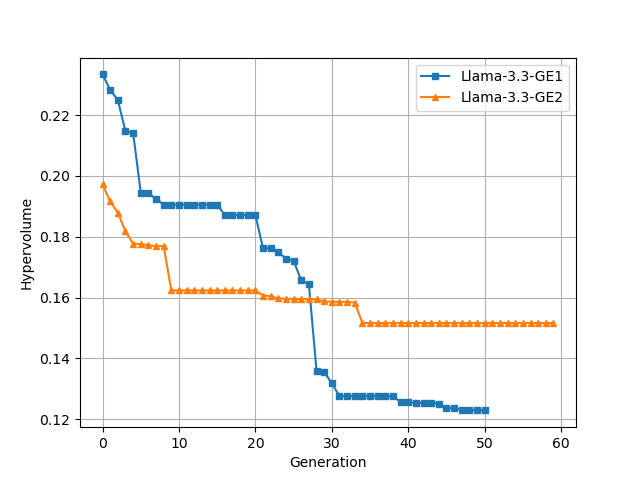}
  \caption{Hypervolume for LLM-Guided Evolution of Llama-3.3-GE1 and  Llama-3.3-GE2.}
  \Description{Hypervolumes per generations for each GEs.}
  \label{fig:hypervolume}
\end{figure}

To compute the hypervolume shown in Figure \ref{fig:hypervolume}, the number of parameters and inference speed are normalized between 0 and 1. Additionally, 1 - precision and 1 - recall are used to align all objectives with a minimization trend. As a result, the scores are contained within a hypercube of unit hypervolume. 

Figure \ref{fig:hypervolume} shows the hypervolumes for Llama-3.3-GE1 and  Llama-3.3-GE2. The blue connected squares denote the hypervolume for Llama-3.3-GE1 at each generation and the orange connected triangles denote the hypervolume for Llama-3.1-GE2 at each generation. The figure shows that Both Llama-3.3-GE1 and Llama-3.3-GE2 improve over time, and that Llama-3.3-GE1 achieves better hypervolume than Llama-3.3-GE2. It is interesting to note that the Llama-3.3-GE1 outperforms the other Llama-GE2 after 25 generations and there was a significant improvement from generation 26 to generation 27. Meanwhile, the hypervolume of Llama-3.3-GE2 reaches a plateau after 33 generations. 

The figure also shows that at 0th generation, the hypervolumes of the two GEs are not the same. This could be caused by the randomness of creating the initial populations or by the difference in the set of seeds. The Llama-3.3-GE1 was only seeded with \ac{yolo}v3 while Llama-3.3-GE2 was seeded with additional seeds of \ac{yolo}v3-tiny,  \ac{yolo}v9s, \ac{yolo}v10-tiny, and \ac{yolo}v11. We note a further investigation on the stochastic nature of this process with significantly more  trials in our proposed future work.

In addition, Llama-3.3-GE1 discovered 135 Pareto-optimal \ac{yolo} variants in its final generation, while Llama-3.3-GE2 generated 128 Pareto-optimal \ac{yolo} variants in its final generation. By inspecting the Pareto front individuals, we observed common changes such as modifications to the number of layers, the width and depth multipliers, and the arguments of the layers (see Listing~\ref{lst:seed_model}), especially in Llama-3.3-GE1.  These observations suggest that Llama-3.3 relies solely on the seed model, \ac{yolo}v3, and lacks awareness of other layer types. In contrast, Llama-3.3-GE2, seeded with multiple state-of-the-art \ac{yolo} models, incorporated a wider variety of layers (See Listings \ref{lst:modified_yaml}-\ref{lst:modified_yaml_3}).

Note that even though Figure \ref{fig:hypervolume} shows that the hypervolume for Llama-3.3-GE2 appears to stagnate around generation 34, Figure \ref{fig:Pareto_frontier_counts} shows that during this time, the evolution is still discovering a significant number of new individuals along the Pareto frontier, suggesting an increase in fidelity along the objectives in the trade-off space. It is a possibility that the Llama-3.3-GE2 is in a state of punctuated equilibrium.

In some cases, the LLM attempted to create new layers as Python code for the configuration. However, our current framework does not yet utilize this information. Future work will focus on integrating these generated layers into the model to further enhance \ac{yolo}'s architecture.


\begin{listing}[ht]
\begin{minted}[fontsize=\scriptsize]{yaml}
# --PROMPT LOG--
# Ultralytics YOLO, AGPL-3.0 license
# YOLO object detection model. For details see https://docs.ultralytics.com/models/yolov3

# --OPTION--
# Parameters
nc: 80 # number of classes
depth_multiple: 1.5 # increased model depth multiple
width_multiple: 1.5 # increased layer channel multiple

# backbone
backbone:
  # [from, number, module, args]
  - [-1, 1, Conv, [40, 3, 1]] # 0, increased channels
  - [-1, 1, Conv, [80, 3, 2]] # 1-P1/2, increased channels
  - [-1, 1, Bottleneck, [80]] # increased channels
  - [-1, 1, Conv, [160, 3, 2]] # 3-P2/4, increased channels
  - [-1, 2, Bottleneck, [160]] # increased channels
  - [-1, 1, Conv, [320, 3, 2]] # 5-P3/8, increased channels
  - [-1, 8, Bottleneck, [320]] # increased channels
  - [-1, 1, Conv, [640, 3, 2]] # 7-P4/16, increased channels
  - [-1, 8, Bottleneck, [640]] # increased channels
  - [-1, 1, Conv, [1280, 3, 2]] # 9-P5/32, increased channels
  - [-1, 4, Bottleneck, [1280]] # 10, increased channels
  - [-1, 1, Bottleneck, [1280]] # Additional bottleneck layer

# head
head:
  - [-1, 1, Bottleneck, [1280, False]]
  - [-1, 1, SPP, [640, [5, 9, 13]]]
  - [-1, 1, Conv, [1280, 3, 1]]
  - [-1, 1, Conv, [640, 1, 1]]
  - [-1, 1, Conv, [1280, 3, 1]] # 15 (P5/32-large)
  - [-1, 1, Conv, [1280, 1, 1]] # Additional convolutional layer

  - [-2, 1, Conv, [320, 1, 1]]
  - [-1, 1, nn.Upsample, [None, 2, "nearest"]]
  - [[-1, 8], 1, Concat, [1]] # cat backbone P4
  - [-1, 1, Bottleneck, [640, False]]
  - [-1, 1, Bottleneck, [640, False]]
  - [-1, 1, Conv, [320, 1, 1]]
  - [-1, 1, Conv, [640, 3, 1]] # 22 (P4/16-medium)

  - [-2, 1, Conv, [160, 1, 1]]
  - [-1, 1, nn.Upsample, [None, 2, "nearest"]]
  - [[-1, 6], 1, Concat, [1]] # cat backbone P3
  - [-1, 1, Bottleneck, [320, False]]
  - [-1, 2, Bottleneck, [320, False]] # 27 (P3/8-small)

  - [[27, 22, 15], 1, Detect, [nc]] # Detect(P3, P4, P5)

\end{minted}
    \caption{Example 1 generated by Llama 3.3-GE2 with mAP@50 of 94.5\%.}
    \label{lst:modified_yaml}
\end{listing}

\begin{listing}[ht]
\begin{minted}[fontsize=\scriptsize]{yaml}

# Parameters
nc: 80 # number of classes
scales: # model compound scaling constants, i.e.'model=yolov10n.yaml' will call yolov10.yaml with scale 'n'
  # [depth, width, max_channels]
  s: [0.6, 0.85, 1280] # Increased depth, width, and max channels for enhanced feature extraction

backbone:
  # [from, repeats, module, args]
  - [-1, 1, Conv, [160, 3, 2]] # 0-P1/2, increased channels for better initial feature extraction
  - [-1, 1, Conv, [320, 3, 2]] # 1-P2/4, increased channels for deeper feature extraction
  - [-1, 5, C2f, [320, True]] # Increased repeats for deeper feature extraction
  - [-1, 1, Conv, [640, 3, 2]] # 3-P3/8
  - [-1, 5, C2f, [640, True]] # Increased repeats for deeper feature extraction
  - [-1, 1, SCDown, [1280, 3, 2]] # 5-P4/16
  - [-1, 5, C2f, [1280, True]] # Increased repeats for deeper feature extraction
  - [-1, 1, SCDown, [1280, 3, 2]] # 7-P5/32, increased channels
  - [-1, 1, SPPF, [1280, 5]] # 8, increased channels
  - [-1, 1, PSA, [1280]] # 9, increased channels

head:
  - [-1, 1, nn.Upsample, [None, 2, "nearest"]]
  - [[-1, 6], 1, Concat, [1]] # cat backbone P4
  - [-1, 5, C2f, [640]] # Increased repeats for enhanced feature fusion

  - [-1, 1, nn.Upsample, [None, 2, "nearest"]]
  - [[-1, 4], 1, Concat, [1]] # cat backbone P3
  - [-1, 5, C2f, [320]] # Increased repeats for enhanced feature fusion

  - [-1, 1, Conv, [640, 3, 2]]
  - [[-1, 12], 1, Concat, [1]] # cat head P4
  - [-1, 5, C2f, [1280]] # Increased repeats for enhanced feature fusion

  - [[12, 15, 18], 1, v10Detect, [nc]] # Detect(P3, P4, P5)


\end{minted}
    \caption{Example 2 generated by Llama 3.3-GE2 .}
    \label{lst:modified_yaml_2}
\end{listing}

\begin{listing}[ht]
\begin{minted}[fontsize=\scriptsize]{yaml}
# Parameters
nc: 80 # number of classes
depth_multiple: 0.5 # model depth multiple
width_multiple: 0.5 # layer channel multiple

# backbone
backbone:
  # [from, number, module, args]
  - [-1, 1, Conv, [32, 3, 1]] # 0
  - [-1, 1, Conv, [64, 3, 2]] # 1-P1/2
  - [-1, 1, Bottleneck, [64]]
  - [-1, 1, Conv, [128, 3, 2]] # 3-P2/4
  - [-1, 1, Bottleneck, [128]]
  - [-1, 1, Conv, [256, 3, 2]] # 5-P3/8
  - [-1, 2, Bottleneck, [256]]
  - [-1, 1, Conv, [512, 3, 2]] # 7-P4/16
  - [-1, 2, Bottleneck, [512]]
  - [-1, 1, SPP, [256, [5, 9, 13]]] # Added SPP module
  - [-1, 1, Conv, [512, 1, 1]] # Added Conv module

# head
head:
  - [-1, 1, Bottleneck, [512, False]]
  - [-1, 1, Conv, [512, 3, 1]]
  - [-1, 1, Conv, [256, 1, 1]]
  - [-1, 1, Conv, [512, 3, 1]] # 10 (P5/32-large)

  - [-2, 1, Conv, [128, 1, 1]]
  - [-1, 1, nn.Upsample, [None, 2, "nearest"]]
  - [[-1, 6], 1, Concat, [1]] # cat backbone P4
  - [-1, 1, Bottleneck, [256, False]]
  - [-1, 1, Conv, [128, 1, 1]]
  - [-1, 1, Conv, [256, 3, 1]] # 16 (P4/16-medium)

  - [-2, 1, Conv, [64, 1, 1]]
  - [-1, 1, nn.Upsample, [None, 2, "nearest"]]
  - [[-1, 3], 1, Concat, [1]] # cat backbone P3
  - [-1, 1, Bottleneck, [128, False]]
  - [-1, 1, Conv, [64, 1, 1]] # 20 (P3/8-small)

  - [[20, 16, 10], 1, Detect, [nc]] # Detect(P3, P4, P5)

\end{minted}
    \caption{Example 3 generated by Llama 3.3-GE2 .}
    \label{lst:modified_yaml_3}
\end{listing}


Listing~\ref{lst:modified_yaml} shows a modified YAML configuration file of the seed model. It was one of the Pareto front individuals proposed by Llama-3.3-GE2.  The modifications made by the LLM-guided mutations and crossovers include changing the inputs for the \textbf{Conv} and \textbf{Bottleneck} layers and changing the number of layers.

\subsection{Performance Improvements}
The application of LLM-Guided Evolution to \ac{yolo} optimization generates promising \ac{yolo} variants with performance improvements across several key metrics, demonstrating the efficacy of this approach in optimizing the architecture and associated hyperparameters when seeded with a state-of-the-art object detection model for a challenging real-world dataset like KITTI. These improvements can be attributed to the framework's ability to intelligently explore the configuration space through LLM-Guided mutations and crossovers. This autonomous framework successfully generated many \ac{yolo} variants, outperforming the original network's performance of 92.5\% mAP@50 on holdout data. A notable example generated by GE with Llama 3.3 achieves 94.5\% mAP@50 on holdout data.


\begin{figure}[h]
  \centering
  \includegraphics[width=\linewidth]{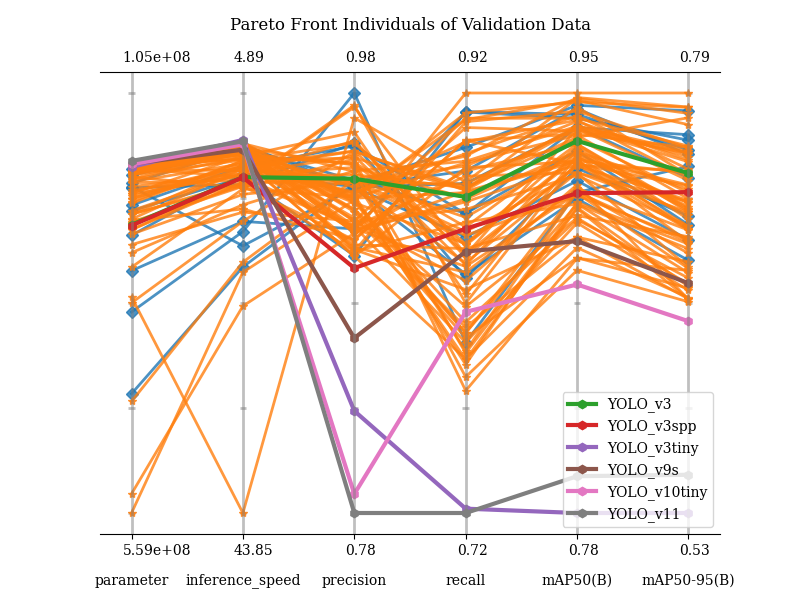}
  \caption{Parallel coordinate plot of overall Pareto front individuals evaluated on the validation Data for both Llama-3.3-GE1 and Llama-3.3-GE2}
  \Description{PCP plot of Pareto frontier with performance of reference models}
  \label{fig:pcp_holdout_val}
\end{figure}
To understand the variety of individuals with unique contributions discovered across both GE1 and GE2, Figure~\ref{fig:pcp_holdout_val} presents a parallel coordinate plot of the individuals on the Pareto frontier evaluated on the validation dataset, which the LLM-Guided Evolution framework uses to assess the performance of newly generated individuals. We combined the 135 Pareto-optimal individuals from GE1 with the 128 Pareto-optimal individuals from GE2, and kept only the individuals that remained Pareto-optimal. These individuals are co-dominant with respect to the objectives of model parameters, inference speed, precision, and recall, while also highlighting two key object detection metrics: mAP@50 and mAP@50-95. In the plot, blue diamonds connected by lines represent the performance of the remaining 11 individuals generated by Llama-3-3-GE1 and orange stars connected by lines indicate the performance of the remaining 63 individuals generated by Llama-3.3 GE2. The figure also shows the performance of \ac{yolo}v3, \ac{yolo}v3 SPP, \ac{yolo}v3 tiny, and other State-of-the-Art \ac{yolo} model for reference. We retrained these models on the same KITTI data with the default settings for training. Optimization of the training setting is not included in this study, but it remains a topic for future exploration.

\begin{figure}[h]
  \centering
  \includegraphics[width=\linewidth]{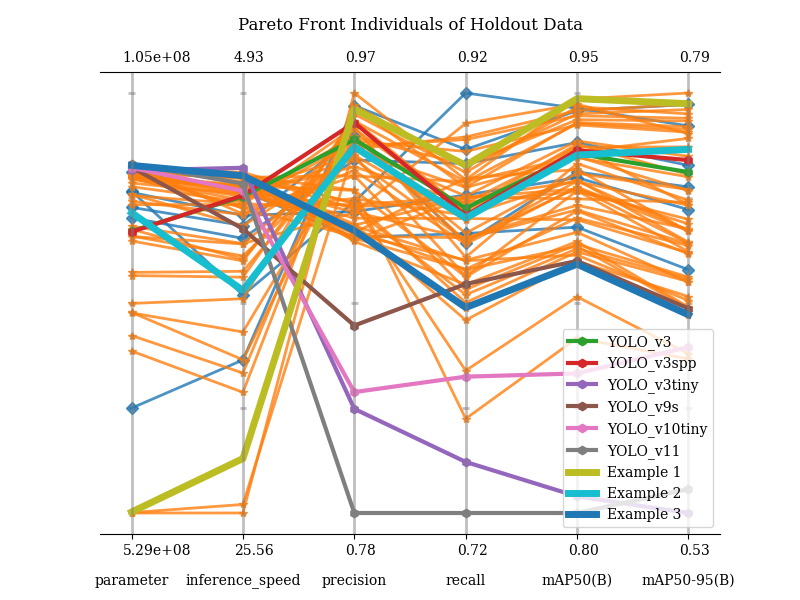}
  \caption{Parallel coordinate plot of overall Pareto front individuals evaluated on the holdout data for both Llama-3.3-GE1 and Llama-3.3-GE2 }
  \Description{PCP of Pareto frontier and performance of reference models}
  \label{fig:pcp_holdout_test}
\end{figure}
Similar to Figure~\ref{fig:pcp_holdout_val}, Figure~\ref{fig:pcp_holdout_test} presents the parallel coordinate plot of the Pareto front individuals evaluated on the holdout dataset. We reduced the set of individuals using the same procedure as applied to the validation set. The figure illustrates that some individuals outperform the reference models, including \ac{yolo}v3, \ac{yolo}v3 SPP, \ac{yolo}v3 tiny, \ac{yolo}v9s, \ac{yolo}v10tiny and \ac{yolo}v11. However, these improved variants come at the cost of higher computational complexity, making them more resource-intensive. In the plot, blue diamonds connected by lines represent the performance of 6 individuals generated by Llama-3.3-GE1 and orange stars connected by lines indicate the performance of 45 individuals generated by Llama-3.3-GE2. Listings \ref{lst:modified_yaml}-\ref{lst:modified_yaml_3} are examples 1-3 highlighted in  Figure~\ref{fig:pcp_holdout_test}. They demonstrate that the LLM-GE is able to explore the configuration space by modifying numbers of layers, adding layers ( such as \textbf{SPP} layer and \textbf{Conv} layer), and changing the inputs of layers. 

These visualizations effectively illustrate that the LLM-GE successfully evolved YOLO variants with trade-off between mAP and speed. This result highlights the flexibility and effectiveness of GE in real-world computer vision challenges, offering a novel paradigm for autonomous model optimization that combines LLM-driven reasoning with evolutionary strategies.

\section{Conclusions and Future Work}
We introduced a novel approach of using a LLM-GE to intelligently modify neural network architectures through their module, layer, and hyperparameter descriptions in YAML configuration files to optimize the \ac{yolo} object detection model on real-world KITTI data. Our approach focuses on modifying the YAML configuration file that defines \ac{yolo} architectures within the Ultralytics codebase, enabling structured and interpretable optimization. This method allows for systematic exploration of hyperparameter spaces and architecture variations in a manner that is both efficient and scalable. Our results are promising, as \ac{llm}-\ac{ge} successfully evolved multiple \ac{yolo} variants with significant improvements in performance over seeded individuals and state-of-the-art implementations. By allowing \acp{llm} to drive the evolutionary process, we observed notable enhancements in object detection accuracy.

This study serves as a proof-of-concept for evolving YAML configuration files using LLM-GE. While our approach has demonstrated clear potential, there remains significant opportunity for further research. Future work should explore more sophisticated evolutionary techniques, including island migration with multiple simultaneous LLMs, co-evolutionary improvements using prompts specifically tailored to LLMs and tasks at hand, and fine-grained parameter tuning. Additionally, expanding the scope of LLM-GE beyond YAML configurations to incorporate automated code generation for implementing new layers in adherence with the YAML settings could further enhance \ac{yolo}’s adaptability. Increasing the number of repeated trials for rigorous statistical analysis and continuing integration with SoTA open-source LLMs, such as DeepSeek-R1, Codestral, and Mixtral 8x22B, will be crucial for further advancements. It is also worth exploring the impact of seeds on the evolutionary process. 

Other areas of future growth involve the manner of integration with ongoing research in the utilization of \acp{llm}, including concepts such as \ac{rag} or fine-tuning to improve the performance of the mating and mutation operations that the \acp{llm} are responsible for.

Ultimately, our goal is to continue to build a framework that adapts easily to new problem domains, objectives, and state-of-the-art models. We will continue to automate the evolution of \ac{yolo} variants to outperform existing SoTA models across all key metrics. As \acp{llm} continue to advance, they will play an increasingly pivotal role in automating and refining neural network design, paving the way for more efficient, adaptive, and high-performing AI models.

\begin{acks}
The authors wish to acknowledge Clint Morris, the original first-author of \ac{llm}-\ac{ge}, and whose flexible codebase enabled the research performed in this paper.
\end{acks}

\bibliographystyle{ACM-Reference-Format}
\bibliography{llm-ge-workshop}


\end{document}